\documentclass{llncs}
\usepackage[T1]{fontenc}
\usepackage[latin9]{inputenc}
\usepackage{array}
\usepackage{booktabs}
\usepackage{algorithm2e}
\usepackage{amstext}
\usepackage{amssymb}
\usepackage{graphicx}

\makeatletter

\providecommand{\tabularnewline}{\\}

\usepackage{microtype}
\usepackage[overload]{textcase}
\usepackage{tikz}
\usetikzlibrary{arrows,shapes}

\makeatother

\begin{document}

\title{Lamarckian Evolution of Convolutional Neural Networks}

\author{Jonas Prellberg \and Oliver Kramer}

\institute{University of Oldenburg, Oldenburg, Germany\\
\email{\textnormal{\{}jonas.prellberg,oliver.kramer\textnormal{\}}@uni-oldenburg.de}}
\maketitle
\begin{abstract}
Convolutional neural networks belong to the most successul image classifiers,
but the adaptation of their network architecture to a particular problem
is computationally expensive. We show that an evolutionary algorithm
saves training time during the network architecture optimization,
if learned network weights are inherited over generations by Lamarckian
evolution. Experiments on typical image datasets show similar or significantly
better test accuracies and improved convergence speeds compared to
two different baselines without weight inheritance. On CIFAR-10 and
CIFAR-100 a 75\,\% improvement in data efficiency is observed.

\end{abstract}

\section{Introduction}

Over the last years, deep neural networks and especially convolutional
neural networks (CNN) have become state-of-the-art in numerous application
domains. Their performance is sensitive to the choice of hyperparameters,
such as learning rate or network architecture, which makes hyperparameter
optimization an important aspect of applying neural networks to new
problems. However, such optimization is computationally expensive
due to the lengthy training process that has to be repeated each time
a new hyperparameter setting is tested. This downside applies to optimization
by hand as well as to automated approaches, such as grid search, random
search or evolutionary optimization.

Earlier neuroevolution works \cite{baluja1996evolution,1199654,Stanley:2009:HEE:1516090.1516093,doi:10.1162/106365602320169811}
optimize network architectures together with the network weights using
an evolutionary algorithm (EA). In recent years though, EAs have mostly
been applied to optimize network hyperparameters, while the training
is performed with backpropagation since it offers increased efficiency
for training the large and deep networks prevalent today. The usual
procedure is as follows: A genotype encodes the network architecture
and corresponding hyperparameters. In a genotype-phenotype process,
a network is built from this description and initialized with random
weights. Then, several epochs of backpropagation on a training set
adjust the network weights. Finally, the network is tested on a validation
set and a metric, such as accuracy, is reported as the genotype's
fitness.

Instead of randomly initializing the network weights before training,
it is also possible to inherit the already learned weights of an ancestor
network. This inheritance of aquired traits is a form of lamarckian
evolution which, while rejected in biology, can prove useful in artificial
evolution. In this work, we show that weight inheritance can drastically
increase the data efficiency of an EA that optimizes neural network
architectures.

The remainder of this paper is organized as follows: Section~\ref{sec:related-work}
presents related work about approaches that optimize architectures
with EAs or use weight inheritance in their EAs. Section~\ref{sec:method}
describes the EA that is used in this paper and explains how the mutation
with weight inheritance works. In Section~\ref{sec:experiments}
experimental results are presented and discussed. The paper ends with
a conclusion in Section~\ref{sec:conclusion}.

\section{Related Work\label{sec:related-work}}

Lamarckian evolution describes the idea that traits aquired over the
lifetime of an individual are inherited to its offspring \cite{Sasaki2000}.
While rejected in biology, this approach can be beneficial for artificial
evolution when there is a bi-directional mapping between genotype
and phenotype. This allows to encode learned behavior back into the
genotype and then apply an EA as usual. For example, Parker and Bryant
\cite{cec-quake2} and Ku et al. \cite{lamarck-rnn} apply lamarckian
evolution to neural networks by directly encoding the network weights
in the genotype. This creates a simple one-to-one mapping between
genotype and phenotype.

NEAT \cite{doi:10.1162/106365602320169811} is a neuroevolution algorithm
that allows to grow neural networks starting with a minimal network
and expanding it through mutation and principled crossover between
the graphs. Because NEAT operates on single graph nodes and edges,
it has been most successful on problems that can be solved with small
neural networks. However, NEAT has inspired many approaches that try
to extend the concept to evolving graphs of higher-level operations,
such as convolutions.

Desell \cite{Desell:2017:LSE:3067695.3076002} uses a variant of NEAT
to optimize the structure of a CNN that trains individual networks
using backpropagation. An experiment with weight inheritance was conducted
but it was not found to decrease the time necessary to train a single
network to completion. Fernando et al. \cite{Fernando:2016:CED:2908812.2908890}
use a microbial genetic algorithm to optimize the structure of a DPPN,
which is a network that produces weights for a new network. Weight
inheritance was found to improve the MSE in an image reconstruction
experiment. Verbancsics and Harguess \cite{mnist-hyperneat} test
HyperNEAT \cite{Stanley:2009:HEE:1516090.1516093} as a way to train
CNNs for image classification. However, results were mediocre and
could be substantially improved using backpropagation.

Kramer \cite{2017arXiv170903247K} uses a $(1+1)-\text{EA}$ to optimize
the hyperparameters defining a convolutional highway network. Suganuma
et al. \cite{Suganuma:2017:GPA:3071178.3071229} use a modified $(1+\lambda)-\text{EA}$
to optimize the structure of a CNN using a Cartesian genetic programming
encoding scheme. Both approaches use small populations and only employ
mutations while still achieving good results.

Weight inheritance is already employed in other works to varying degrees
and for various reasons. For example, Jaderberg et al. \cite{2017arXiv171109846J}
use an EA to optimize hyperparameters of static networks. It only
performs mutations but inherits weights to mutated offspring. The
networks are therefore trained to completion over multiple steps with
potentially different hyperparameters. If one of the optimized hyperparameters
is the learning rate, this effectively trains the network using a
dynamic, evolved learning rate schedule. Instead, our goal is to highlight
the data efficiency gains that come with weight inheritance.

Real et al. \cite{2017arXiv170301041R} use an EA with a very large
population size and an unprecedented amount of computational resources
to optimize the structure of a CNN. The method uses only mutation
and inherits trained network weights through mutation. Training with
backpropagation is performed for about 28 epochs per fitness evaluation
on CIFAR-10 and CIFAR-100 and competitive results are reached. Furthermore,
weight inheritance is shown to improve the test accuracy that the
final network achieves. While their approach is similar to our work,
we strive to keep computational demands low with the goal to reduce
requirements further with the inclusion of weight inheritance. 

Unrelated to evolutionary methods, the Net2Net algorithm \cite{net2net}
is an interesting approach to accelerate the sequential training of
multiple related models. Starting from a trained model, it is possible
to increase its depth or width while keeping the represented function
the same. This is achieved by choosing the new weights in such a way
that their effects cancel out. The training of the new model then
progresses faster because the initial weights are already very good.

\section{Method\label{sec:method}}

To assess the influence of weight inheritance for neural network architecture
optimization, design decisions regarding the optimizable hyperparameters
and type of EA must be made. The goal is not to achieve state-of-the-art
performance or find novel architectures but instead to show the advantages
of weight inheritance. Therefore, we choose to optimize a fairly restricted
architecture space which, however, is still applicable to many problems.
This allows the EA to converge fast enough within our hardware resource
constraints to make multiple repetitions of the same experiment feasible
for statistical purposes.

\begin{figure}
\centering
\scriptsize
\definecolor{niceblue}{RGB}{0, 88, 242}
\begin{tikzpicture}[
    scale=0.7,
    every node/.style={transform shape},
	node distance=8mm,
    >=stealth',
    bend angle=45,
    auto,
	box/.style={
		draw,inner sep=1.2mm,
		minimum width=4.2cm, minimum height=4.5mm
	},
]

\node [box] (img) {Image};
\node [box, fill=niceblue!30] (c1) [below of=img] {$\textnormal{Conv2D}\left(f_1, k_1, s_1\right), \textnormal{BN}, \textnormal{ReLU}$};
\node [] (dots) [below of=c1] {$\cdots$};
\node [box, fill=niceblue!30] (ci) [below of=dots] {$\textnormal{Conv2D}\left(f_i, k_i, s_i\right), \textnormal{BN}, \textnormal{ReLU}$};
\node [box] (pool) [below of=ci] {Global Average Pooling};
\node [box] (dense) [below of=pool] {Dense, Softmax};

\draw[->] (img) to (c1);
\draw[->] (c1) to (dots);
\draw[->] (dots) to (ci);
\draw[->] (ci) to (pool);
\draw[->] (pool) to (dense);

\end{tikzpicture}

\caption{Graph template defining the network architecture search space. Each
building block (shown in blue) has individual hyperparameters filter
count $f_{i}$, kernel size $k_{i}$ and stride $s_{i}$ that have
to be optimized by the EA. \label{fig:graph-template}}
\end{figure}
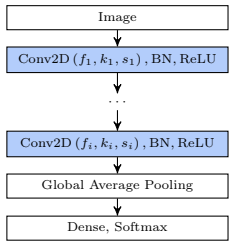

The architecture search space is defined by the template presented
in Figure~\ref{fig:graph-template}. It is made from stacked building
blocks that consist of a convolutional layer followed by batch normalization
\cite{ioffe2015batch} and a ReLU activation. The number of building
blocks and the individual number of filters, kernel size and stride
of the convolutional layer in each building block are subject to optimization.
We statically append global average-pooling, a dense layer and a softmax
activation function after the last building block, since experiments
are performed specifically on image datasets.

The optimization is performed by a $(1+1)-\text{EA}$. In contrast
to evolutionary algorithms with larger populations, the necessary
computational resources are modest, but the method is also more prone
to getting stuck in local optima in multi-modal problems. To alleviate
this, a form of niching is introduced. The evolutionary algorithm
and its mutation operator are presented in more detail in the following
sections.

\subsection{Evolutionary algorithm}

Algorithm \ref{alg:opo-ga} presents the $(1+1)-\text{EA}$ with niching
as pseudo-code. An initial network consisting of a single convolutional
layer with random filter count, random kernel size and a stride of
one is created. This parent network is optimized by the EA as follows:
First, a random mutation from the set of possible mutations is applied
to the parent to create a child network. Next, the child network's
fitness is evaluated. This means the network is trained for $e$ epochs
and the validation set accuracy is returned as its fitness. If the
child's fitness is greater than the parent's fitness, the child replaces
the parent.

Because this algorithm is greedy, it can get stuck in local minima.
Therefore, a niching approach adapted from Kramer \cite{2017arXiv170903247K}
is implemented. There is a random chance $\eta$ to follow solutions
that are initially worse. In such a case, a child, which has a lower
fitness than its parent, is used as the parent network for a recursive
call of the same algorithm. During niching, the mutate-evaluate-select-loop
is repeated $k$ times. When the last loop iteration ends, the best
network found during niching is returned. If this network has a greater
fitness than the original parent, it is selected. Otherwise, optimization
proceeds with the original parent.

\begin{algorithm}
\DontPrintSemicolon
\SetProgSty{}

$a \leftarrow$ initial network\;
\While{termination condition not met}{
  $b \leftarrow \textnormal{mutate}\left(a\right)$\;
  \uIf{$\textnormal{fitness}\left(b\right)>\textnormal{fitness}\left(a\right)$}{
    $a\leftarrow b$\;
  }
  \ElseIf{$\textnormal{random}()<\eta$ and not yet niching}{
    $c\leftarrow$ best network after recursion with $b$ as initial network for $k$ iterations (niching)\;
    \If{$\textnormal{fitness}\left(c\right)>\textnormal{fitness}\left(a\right)$}{
      $a\leftarrow c$\;
    }
  }
}
\Return{$a$}\;

\vspace{2mm}

\caption{$(1+1)-\text{EA}$ with niching\label{alg:opo-ga}}
\end{algorithm}

\subsection{Mutation operator}

As mentioned before, the number of building blocks and the number
of filters, kernel size and stride of each convolutional layer are
subject to  optimization. For simplicity, all these hyperparameters
are chosen from predefined sets:
\begin{itemize}
\item Number of building blocks in $\mathbb{N}$
\item Filter counts in $\mathcal{F}=\left\{ 16,32,64,96,128,192,256\right\} $
\item Kernel size in $\mathcal{K}=\left\{ 1,3,5\right\} $
\item Stride in $\mathcal{S}=\left\{ 1,2\right\} $
\end{itemize}
Mutations are picked randomly from the list below. Each choice has
a relative frequency (indicated by the multiplier in front of the
list item) that determines how much more likely it is to be chosen
than the mutation with a relative frequency of one. The frequencies
have been chosen such that the more granular mutations, which are
likely to have a smaller impact on the result, are applied less often
in order to effectively use the available computation time.
\begin{itemize}
\item $3\,\times\,$\emph{add block}: Adds a building block at a random
position. The contained convolutional layer is initialized with a
random filter count, random kernel size and a stride of one.
\item $3\,\times\,$\emph{remove block}: Removes a random building block.
\item $2\,\times\,$\emph{add filters}: Picks a random convolution and sets
its filter count to the next greater value in $\mathcal{F}$.
\item $2\,\times\,$\emph{remove filters}: Picks a random convolution and
sets its filter count to the next lower value in $\mathcal{F}$.
\item $2\,\times\,$\emph{change kernel size}: Picks a random convolution
and randomly draws its kernel size from $\mathcal{K}$.
\item $1\,\times\,$\emph{change stride}: Picks a random convolution and
randomly draws its stride from $\mathcal{S}$.
\end{itemize}
All random choices within each mutation, such as picking a random
convolution or kernel size, are drawn uniformly at random from the
appropriate set.

The mutation operator is forced to modify the network. A history of
all previously evaluated networks is maintained and mutations are
repeatedly applied to the parent network until a network is created
that has not been evaluated before. Furthermore, only networks with
at most three convolutions of stride two are allowed because the image
inputs of CIFAR are only $32\times32$ and each stride-two convolution
halves the side lengths.

\subsection{Weight inheritance}

Each network is associated with a set of weights that contains, for
example, kernels and biases for convolutional layers. When creating
the initial parent network, these weights are randomly initialized
in an appropriate fashion, e.g. Glorot \cite{pmlr-v9-glorot10a} initialization.
However, once a network has been evalutated its weights contain useful,
learned values. When the mutation operator is applied, most of these
weights are kept intact. The mutation \emph{change stride} does not influence any
existing weights so that all of them can be reused. The mutation \emph{change kernel size}
affects only the weights of the mutated block which are randomly reinitialized.
However, since the shape of a convolutional layer's kernel depends on its input and output
shape, changing the number of filters in a block affects the block itself
and also the successive block. Therefore, the mutations \emph{add block}, \emph{remove block},
\emph{add filters}, \emph{remove filters} also randomly reinitialize the weights of the
following block if the shapes are incompatible. The additional weights that belong
to the convolutional and batch-normalization layers created by \emph{add
block} are randomly initialized as well.

\section{Experiments\label{sec:experiments}}

Training neural networks for image classification typically takes
lots of resources. Hence, improving data-efficiency would be of great
value. Therefore, we choose to experiment on the standard image benchmarks
CIFAR-10 and CIFAR-100.

\subsection{Setup\label{subsec:setup}}

The mutation operator that employs weight inheritance is compared
to a mutation operator that randomly reinitializes all network weights
after each mutation. Otherwise, the same EA with the same hyperparameters
is used on both datasets. The niching rate and depth are set to $\eta=0.1$
and $k=5$ respectively.

During each fitness evaluation, a network is trained for $e$ epochs
and subsequently its performance is assessed on the validation set.
Choosing $e$ is a trade-off between evaluation speed and the accuracy
of the fitness assessment. If $e$ is very low, evaluation is fast
but networks are not trained to completion. Therefore the reported
fitness will usually be lower than what the network could actually
achieve given enough training time. Consequently, large but accurate
networks have difficulty competing with smaller networks which train
faster but might reach a lower final accuracy. If $e$ is very high,
these problems vanish but the evaluation takes a long time. Since
many evaluations are necessary for large search spaces, this is impractical.
Weight inheritance is supposed to offset some of the problems that
come with small choices of $e$. The EA gets a budget of $n$ total
training epochs as its termination condition. This allows for a comparison
of accuracy in terms of training examples that each experiment has
processed. Also, the choices of $n$ and $e$ together influence how
many generations, i.e. mutations, are possible within the total training
epoch budget. 

We propose an EA with weight inheritance and $e=4$ training epochs
per fitness evaluation. Note that 4 epochs is not sufficient to train
networks that work well on CIFAR to completion. The comparison baseline
is an EA that does not use weight inheritance with two different epoch
settings. The first baseline, which will be called baseline I, also
uses 4 training epochs per evaluation in order to allow for a direct
comparison. This allows us to show that the algorithm with weight
inheritance is more data efficient and has better final accuracy all
else being equal. The second baseline, which will be referred to as
baseline II, uses $e=16$ training epochs per evaluation. This significantly
longer training time is more in line with the traditional approach
of optimizing neural network architectures. It allows us to show that
our observations still hold here and we do not simply trade a lower
final accuracy for data efficiency. During evolution the best network
is saved in regular intervals. After the EA finishes, these checkpoints
are trained to completion and evaluated on the test set. We use this
to compare test accuracies at one point during the evolution and after
the evolution is finished.

The experiments are repeated 20 times with different random seeds
to account for variance introduced by the randomness that is inherent
to the EA and also the network training. Using the results from these
repetitions, we perform significance tests using the one-sided Mann-Whitney
U test. It was chosen because the sample sizes are small as each sample
requires considerable time to create.

\subsection{Training details}

The datasets have been split into a training, validation, and test
set which contain 45k, 5k and 10k examples respectively for both CIFAR-10
and CIFAR-100. During a fitness evaluation, backpropagation is performed
on the training set for $e$ epochs and the validation set accuracy
is reported as the network's fitness. All training phases are performed
using a cross-entropy loss, the Adam~\cite{adam} optimizer, a batch
size of 512 and a learning rate of $0.001$. Adam's state, i.e. first
and second moment estimates, is not inherited during mutation. The
test set is only used after all experiments have finished to evaluate
saved network checkpoints. These checkpoints are trained to completion
using a learning rate schedule: $10^{-3}$ until epoch 10, $10^{-4}$
until epoch 20 and $10^{-5}$ until epoch 30.

\subsection{Results}

\begin{figure}
\begin{centering}
\includegraphics[width=0.49\columnwidth]{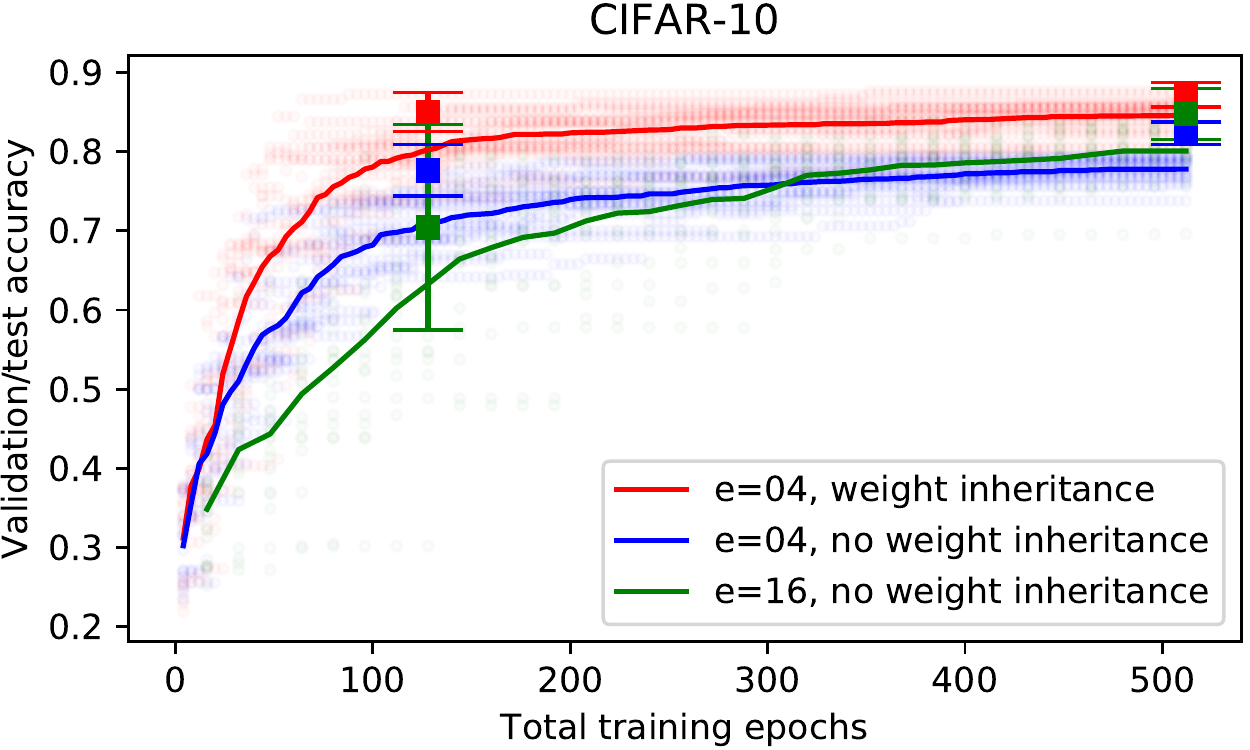}\hfill{}\includegraphics[width=0.49\columnwidth]{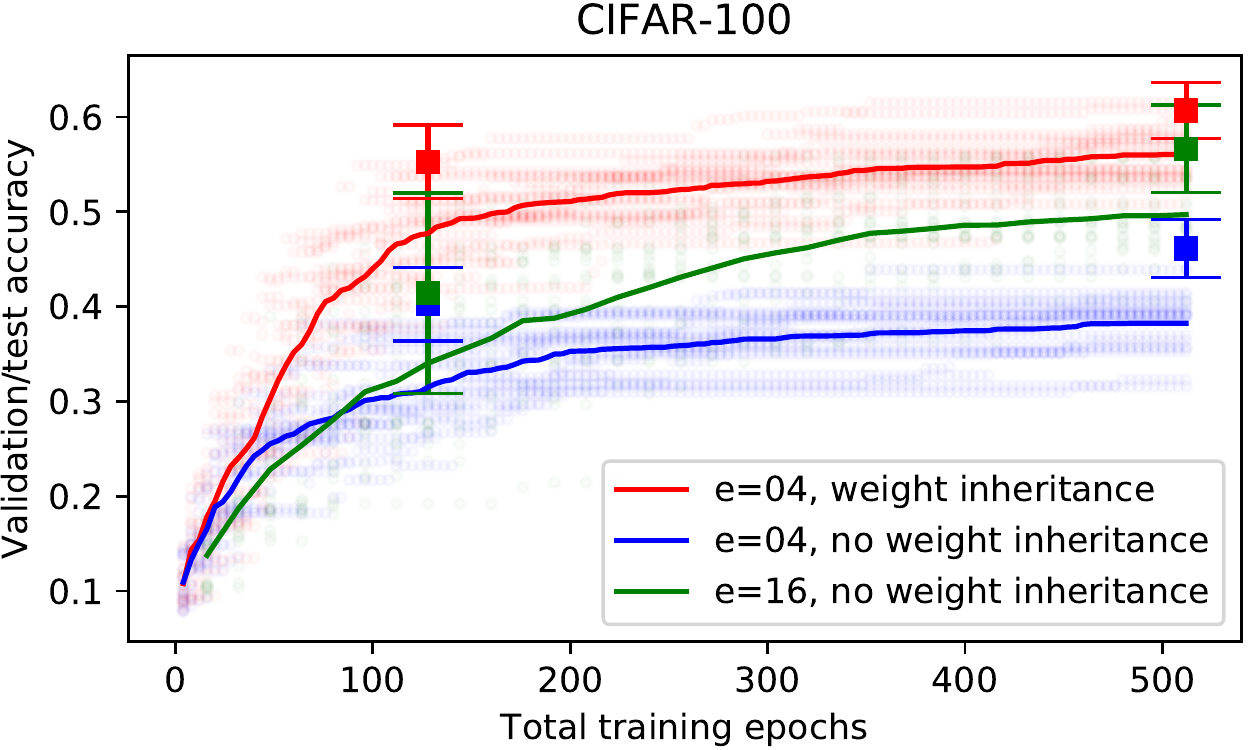}
\par\end{centering}
\caption{Comparison of the EA with weight inheritance and $e=4$ epochs against
two baselines without weight inheritance and $e\in\text{\ensuremath{\left\{  4,16\right\} } }$
epochs on CIFAR-10 and CIFAR-100. Each dot represents the best validation
accuracy achieved so far during an EA run at the respective epoch.
Each line runs through the mean of the dots that are from the same
experimental setting after the same amount of total epochs. Each box
shows the average test accuracy after training the networks to convergence
and the boxplot whiskers represent one standard deviation. \label{fig:cifar-10-100}}
\end{figure}

Figure \ref{fig:cifar-10-100} compares three experimental settings
on CIFAR-10 and CIFAR-100 with a total epoch budget of 512. The EA
with weight inheritance outperforms the comparison baselines that
do not use weight inheritance on both datasets. The accuracy plateau
is reached more quickly and higher test accuracy is achieved.

For CIFAR-10, weight inheritance experiments reach a mean test accuracy
of $85\,\%\pm\,2\%$ after only 128 total training epochs. In comparison,
baseline I experiments reach a mean test accuracy of $82\,\%\pm1\%$
after 512 epochs. This means that the EA with weight inheritance achieves
significantly $\left(p<0.01\right)$ higher accuracy than baseline
I in 75\,\% less total training epochs. Baseline II experiments reach
a test accuracy of $85\,\%\pm3\%$ after 512 epochs. This is slightly,
though not significantly, lower than the test accuracy of the weight
inheritance experiments. After 512 epochs, the weight inheritance
experiments reach a mean test accuracy of $87\,\%\pm2\,\%$ which
is significantly $\left(p<0.01\right)$ higher than baseline II at
512 epochs.

For CIFAR-100 results look very similar. After 128 epochs, the weight
inheritance experiments achieve a mean test accuracy of $55\,\%\pm4\,\%$.
In contrast, baseline I experiments reach a significantly $\left(p<0.01\right)$
lower mean test accuracy of $46\,\%\pm3\,\%$ after 512 epochs. Again,
this is an improvement using 75\,\% less total training epochs. Baseline
II experiments achieve a (not significantly) higher mean test accuracy
of $57\,\%\pm5\,\%$ after 512 epochs. Running the weight inheritance
experiments for all 512 epochs as well results in a mean test accuracy
of $61\,\%\pm3\,\%$ which now is significantly $\left(p<0.01\right)$
higher than baseline II at 512 epochs.

In summary, weight inheritance experiments on CIFAR-10 and CIFAR-100
have shown to achieve significantly $\left(p<0.01\right)$ higher
accuracy using a quarter of the total training epochs when compared
to baseline I that uses the same amount of training epochs per fitness
evaluation. Furthermore, final accuracy after 512 epochs is also significantly
$\left(p<0.01\right)$ higher compared to baseline II experiments
which benefited from more training epochs per fitness evaluation.

\begin{figure}
\begin{centering}
\includegraphics[width=0.5\textwidth]{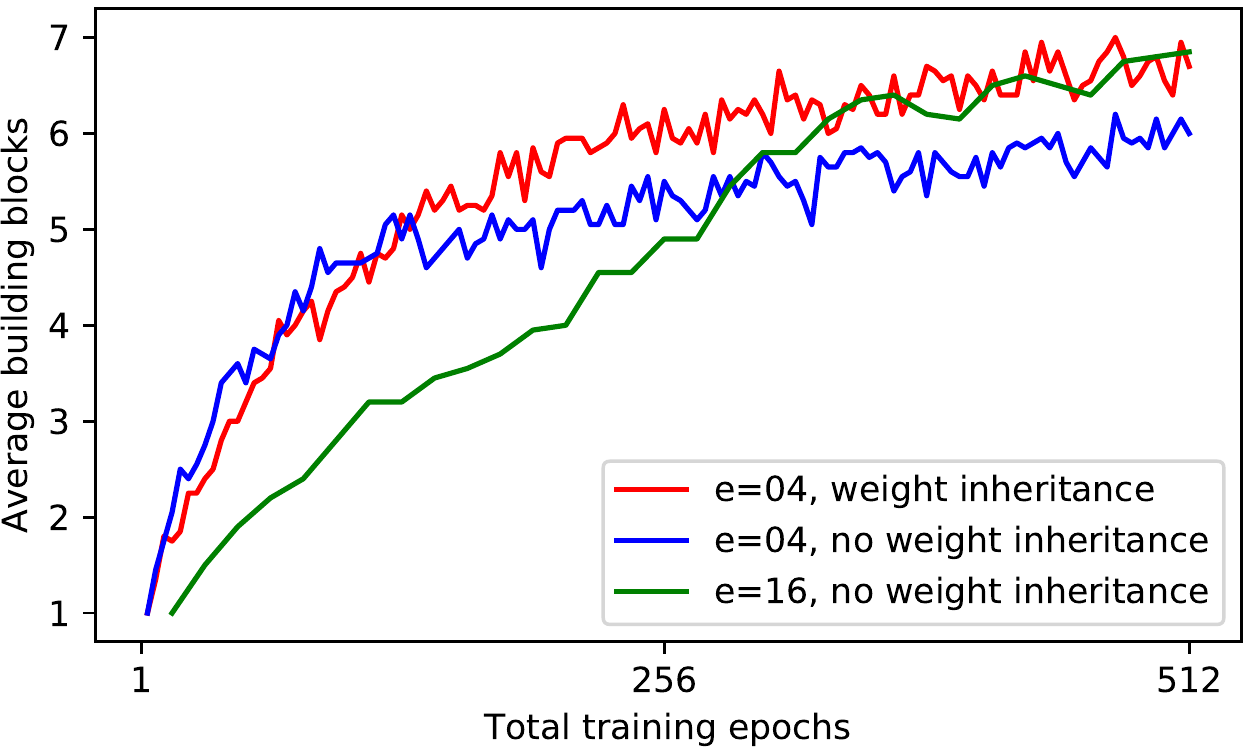}
\par\end{centering}
\caption{Average (of all EA runs) number of building blocks in the genome during
the optimization process on CIFAR-100 \label{fig:cifar100-layers-comparison}}

\end{figure}

To get an idea how the evolutionary process modifies the genotypes,
consider Figure~\ref{fig:cifar100-layers-comparison}. It shows how
the genome length, i.e. the number of building blocks in the corresponding
networks, changes over the course of evolution. All EA runs are initialized
with a genotype that contains a single building block. During the
evolutionary process, increasingly larger genotypes are evaluated
as they offer more accuracy than genotypes with fewer building blocks.
The weight inheritance experiments and baseline II both settle around
an average of 7 building blocks, whereas baseline I networks contain
an average of 6 building blocks.

Table \ref{tab:test-accuracies} lists minimum, mean and maximum test
accuracies of the CIFAR experiments for specific checkpoint epochs.
When weight inheritance is used, minimum, mean and maximum accuracies
are higher than their baseline counterparts at all tested checkpoints.
None of the results reach state-of-the-art performance, which was,
as already pointed out, not the goal of this work. Our best evolved
network on CIFAR-100 without data augmentation reaches a test accuracy
of 66.1\,\% after 512 total epochs and required $6\times10^{10}$
FLOPS\footnote{The FLOPS estimate for a single network is based on the FLOPS reported
by the TensorFlow profiler to process a single example multiplied
by 4 epochs, 98 batches per epoch and batch size 512. The total FLOPS
of the EA run is the sum of the FLOPS estimates for all networks that
were trained during the optimization.} to find. This takes about 1.5 days on a single Nvidia K40 GPU.

\begin{table}
\setlength{\tabcolsep}{2pt}
\renewcommand{\arraystretch}{1.1}
\begin{centering}
\begin{tabular}{ccc>{\centering}p{4mm}ccc>{\centering}p{4mm}ccc}
\multicolumn{3}{c}{Settings} &  & \multicolumn{3}{c}{128 total epochs} &  & \multicolumn{3}{c}{512 total epochs}\tabularnewline
\midrule 
Data & Inheritance & $e$ &  & min & mean & max &  & min & mean & max\tabularnewline
\midrule
C10 & Yes & 4 &  & \textbf{79.1} & \textbf{85.0$\pm$2.4} & \textbf{89.0} &  & \textbf{83.3} & \textbf{87.2$\pm$1.5} & \textbf{89.3}\tabularnewline
 & No & 4 &  & 68.3 & 77.6$\pm$3.2 & 81.8 &  & 78.9 & 82.3$\pm$1.4 & 84.2\tabularnewline
 & No & 16 &  & 32.5 & 70.4$\pm$12.6 & 85.5 &  & 76.6 & 84.8$\pm$3.1 & 88.9\tabularnewline
\midrule
C100 & Yes & 4 &  & \textbf{47.7} & \textbf{55.3$\pm$3.8} & \textbf{61.1} &  & \textbf{56.1} & \textbf{60.7$\pm$2.9} & \textbf{66.1}\tabularnewline
 & No & 4 &  & 31.7 & 40.2$\pm$3.8 & 46.0 &  & 39.8 & 46.1$\pm$3.0 & 52.2\tabularnewline
 & No & 16 &  & 25.9 & 41.4$\pm$10.3 & 57.7 &  & 46.6 & 56.7$\pm$4.5 & 63.0\tabularnewline
\end{tabular}
\par\end{centering}
\vspace{3mm}

\caption{Test accuracies at the two checkpoints on CIFAR-10 and CIFAR-100 \label{tab:test-accuracies}}
\end{table}

\begin{figure}
\noindent \begin{centering}
\includegraphics[width=0.49\columnwidth]{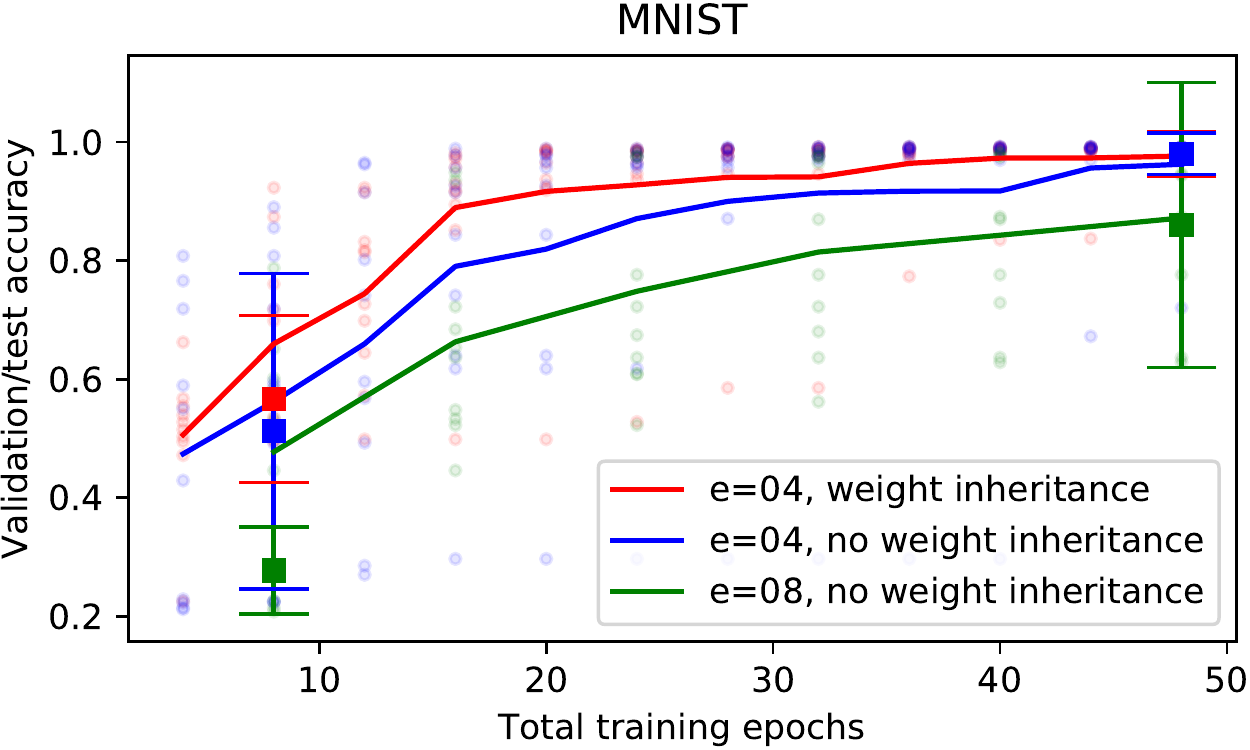}\hfill{}\includegraphics[width=0.49\columnwidth]{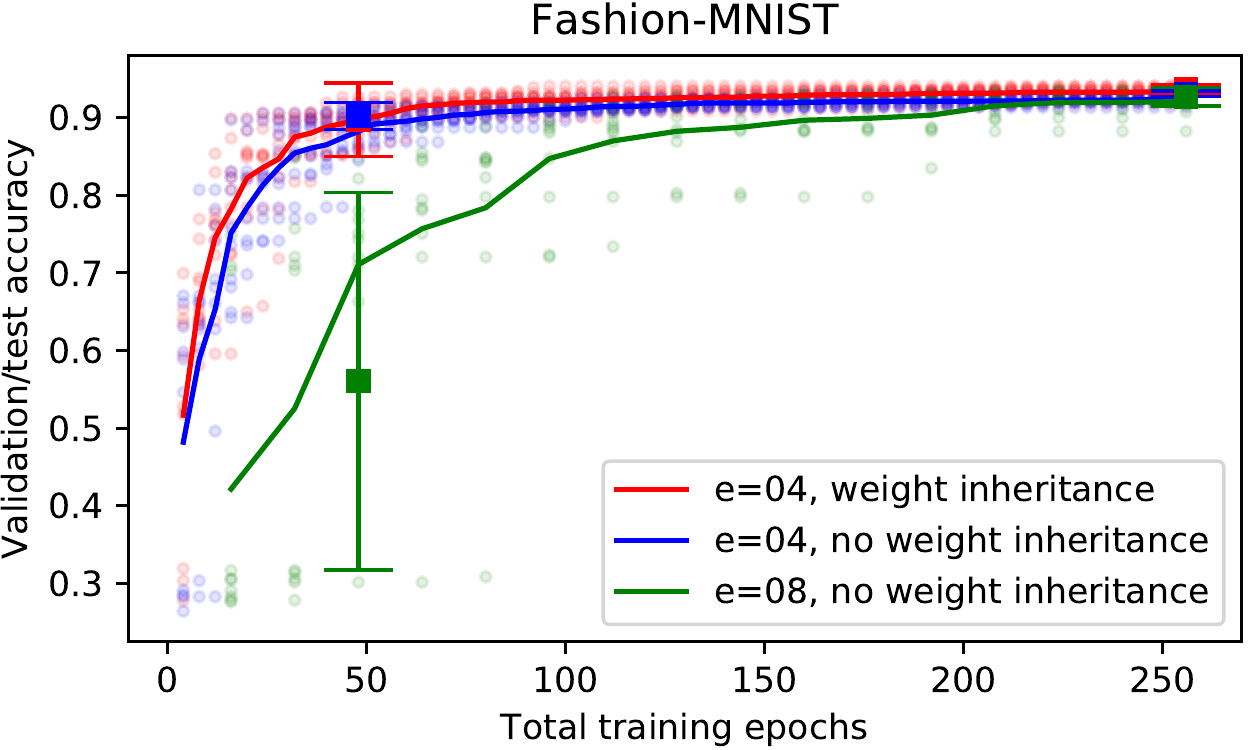}
\par\end{centering}
\caption{Comparison of the EA with weight inheritance and $e=4$ epochs against
two baselines without weight inheritance and $e\in\text{\ensuremath{\left\{  4,8\right\} } }$
epochs on MNIST and Fashion-MNIST. See Figure~\ref{fig:cifar-10-100}
for an explanation of the plot. \label{fig:MNIST-Fashion-MNIST}}
\end{figure}

Additional experiments with 10 repetitions each have been performed
on the MNIST and Fashion-MNIST datasets. The results are shown in
Figure~\ref{fig:MNIST-Fashion-MNIST}. On both datasets, improvements
from weight inheritance over its baselines are marginal. This is expected,
as both datasets are easy to solve compared to CIFAR and can be learned
quickly by small networks. Still, there is no deterioration in performance
from using weight inheritance either.

\subsection{Discussion}

The tradeoff between few and many training epochs per fitness evaluation
that is explained in Section~\ref{subsec:setup} has a visible effect
in Figure~\ref{fig:cifar-10-100}. At the beginning of each experiment,
baseline I outperforms baseline II but at some point this relationship
inverts. This is because small networks, which require only few epochs
to reach good accuracy, are still sufficient to increase the validation
set accuracy in the beginning of the experiment. However, at some
point larger networks become necessary to further improve the results.
These networks require more training time, making it easier for the
algorithm that trains networks longer during fitness evaluation to
progress. Therefore the green and blue graphs intersect. This happens
ealier for CIFAR-100 because it is a harder problem than CIFAR-10.

We have seen weight inheritance experiments consistently outperform
their baselines on CIFAR-10 and CIFAR-100 but could not observe a
significant difference on the MNIST or Fashion-MNIST datasets. We
did not find any instances of our experiments where weight inheritance
was harmful, but this need not be the case generally: Just like in
our work, most recent neuroevolution publications only use mutation
operators and refrain from performing crossover. While this is usually
motivated by the difficulty of designing a useful network crossover
operator, crossover might also bring problems with regard to weight
inheritance. Similarly to choosing a bad initialization before starting
the training of a network, building a new network from trained parts
of different networks could leave it in a region of the parameter
space that is hard to optimize.

\section{Conclusion\label{sec:conclusion}}

Evolutionary algorithms show promise as a way to automatically discover
appropriate network architectures for new problems, but their usefulness
is limited by their enormous computational requirements. Optimizing
deep neural network architectures is computationally expensive because
networks have to be retrained for each fitness evaluation. Therefore,
approaches to lower these requirements are of great value.

We show that an evolutionary algorithm with a weight inheritance scheme
generally achieves equal or higher accuracy compared to baselines
that do not use weight inheritance and benefit from more training
epochs per fitness evaluation. The fitness convergence speed is improved,
sometimes making it possible to drastically reduce the number of total
training epochs, while achieving test accuracies comparable to the
baselines. Specifically, on both CIFAR-10 and CIFAR-100 weight inheritance
increases data efficiency by 75\,\% with comparable test accuracy.
The resulting speedup makes evolutionary algorithms a lot more viable
for application to neural network architecture optimization even on
hard problems. If accuracy is more important than training time, weight
inheritance can also lead to a higher final test accuracy in some
cases. Most importantly though, there has been no instance where weight
inheritance was harmful. All results show either equally good or better
results than the baselines. Thus it seems generally advisable to try
the inclusion of weight inheritance schemes when mutation operators
are used for neural network architecture optimization.

A promising research direction for future work will be to explore
the interactions between weight inheritance and crossover operators.
Also, further decreasing the time necessary for each training step
in the evolutionary process is an important goal. For example, integrating
the Net2Net \cite{net2net} algorithm with the evolutionary algorithm
might offer better results than randomly initializing additional new
weights and allow for even less training steps.

\bibliographystyle{splncs04}
\bibliography{lmkevo}

\begin{thebibliography}{10}
\providecommand{\url}[1]{\texttt{#1}}
\providecommand{\urlprefix}{URL }
\providecommand{\doi}[1]{https://doi.org/#1}

\bibitem{baluja1996evolution}
Baluja, S.: Evolution of an artificial neural network based autonomous land
  vehicle controller. IEEE Transactions on Systems, Man, and Cybernetics, Part
  B (Cybernetics)  \textbf{26}(3),  450--463 (1996)

\bibitem{net2net}
Chen, T., Goodfellow, I., Shlens, J.: Net2net: Accelerating learning via
  knowledge transfer. Proceedings of the International Conference on Learning
  Representations (ICLR '16)  (2016)

\bibitem{Desell:2017:LSE:3067695.3076002}
Desell, T.: Large scale evolution of convolutional neural networks using
  volunteer computing. In: Proceedings of the Genetic and Evolutionary
  Computation Conference Companion (GECCO '17). pp. 127--128. GECCO '17, ACM,
  New York, NY, USA (2017). \doi{10.1145/3067695.3076002},
  \url{http://doi.acm.org/10.1145/3067695.3076002}

\bibitem{Fernando:2016:CED:2908812.2908890}
Fernando, C., Banarse, D., Reynolds, M., Besse, F., Pfau, D., Jaderberg, M.,
  Lanctot, M., Wierstra, D.: Convolution by evolution: Differentiable pattern
  producing networks. In: Proceedings of the Genetic and Evolutionary
  Computation Conference (GECCO '16). pp. 109--116. GECCO '16, ACM, New York,
  NY, USA (2016). \doi{10.1145/2908812.2908890},
  \url{http://doi.acm.org/10.1145/2908812.2908890}

\bibitem{1199654}
Garcia-Pedrajas, N., Hervas-Martinez, C., Munoz-Perez, J.: Covnet: a
  cooperative coevolutionary model for evolving artificial neural networks.
  IEEE Transactions on Neural Networks  \textbf{14}(3),  575--596 (May 2003).
  \doi{10.1109/TNN.2003.810618}

\bibitem{pmlr-v9-glorot10a}
Glorot, X., Bengio, Y.: Understanding the difficulty of training deep
  feedforward neural networks. In: Teh, Y.W., Titterington, M. (eds.)
  Proceedings of the Thirteenth International Conference on Artificial
  Intelligence and Statistics. Proceedings of Machine Learning Research,
  vol.~9, pp. 249--256. PMLR, Chia Laguna Resort, Sardinia, Italy (13--15 May
  2010), \url{http://proceedings.mlr.press/v9/glorot10a.html}

\bibitem{ioffe2015batch}
Ioffe, S., Szegedy, C.: Batch normalization: Accelerating deep network training
  by reducing internal covariate shift. In: Proceedings of the 32nd
  International Conference on Machine Learning (ICML '15). pp. 448--456. Lille,
  France (2015)

\bibitem{2017arXiv171109846J}
{Jaderberg}, M., {Dalibard}, V., {Osindero}, S., {Czarnecki}, W.M., {Donahue},
  J., {Razavi}, A., {Vinyals}, O., {Green}, T., {Dunning}, I., {Simonyan}, K.,
  {Fernando}, C., {Kavukcuoglu}, K.: {Population Based Training of Neural
  Networks}. ArXiv e-prints  (Nov 2017)

\bibitem{adam}
{Kingma}, D.P., {Ba}, J.: {Adam: A Method for Stochastic Optimization}. The
  International Conference on Learning Representations (ICLR '15)  (Dec 2015)

\bibitem{2017arXiv170903247K}
{Kramer}, O.: {Evolution of Convolutional Highway Networks}. In: Applications
  of Evolutionary Computation. Springer International Publishing (2018)

\bibitem{lamarck-rnn}
Ku, K.W.C., Mak, M.W., Siu, W.C.: A study of the lamarckian evolution of
  recurrent neural networks. IEEE Transactions on Evolutionary Computation
  \textbf{4}(1),  31--42 (Apr 2000). \doi{10.1109/4235.843493}

\bibitem{cec-quake2}
Parker, M., Bryant, B.D.: Lamarckian neuroevolution for visual control in the
  quake ii environment. In: 2009 IEEE Congress on Evolutionary Computation. pp.
  2630--2637 (May 2009). \doi{10.1109/CEC.2009.4983272}

\bibitem{2017arXiv170301041R}
Real, E., Moore, S., Selle, A., Saxena, S., Suematsu, Y.L., Le, Q., Kurakin,
  A.: Large-scale evolution of image classifiers. In: Proceedings of the 34th
  International Conference on Machine Learning (ICML '17) (2017),
  \url{https://arxiv.org/abs/1703.01041}

\bibitem{Sasaki2000}
Sasaki, T., Tokoro, M.: Comparison between lamarckian and darwinian evolution
  on a model using neural networks and genetic algorithms. Knowledge and
  Information Systems  \textbf{2}(2),  201--222 (Jun 2000).
  \doi{10.1007/s101150050011}, \url{https://doi.org/10.1007/s101150050011}

\bibitem{Stanley:2009:HEE:1516090.1516093}
Stanley, K.O., D'Ambrosio, D.B., Gauci, J.: A hypercube-based encoding for
  evolving large-scale neural networks. Artif. Life  \textbf{15}(2),  185--212
  (Apr 2009). \doi{10.1162/artl.2009.15.2.15202},
  \url{http://dx.doi.org/10.1162/artl.2009.15.2.15202}

\bibitem{doi:10.1162/106365602320169811}
Stanley, K.O., Miikkulainen, R.: Evolving neural networks through augmenting
  topologies. Evolutionary Computation  \textbf{10}(2),  99--127 (2002).
  \doi{10.1162/106365602320169811}

\bibitem{Suganuma:2017:GPA:3071178.3071229}
Suganuma, M., Shirakawa, S., Nagao, T.: A genetic programming approach to
  designing convolutional neural network architectures. In: Proceedings of the
  Genetic and Evolutionary Computation Conference (GECCO '17). pp. 497--504.
  ACM, New York, NY, USA (2017). \doi{10.1145/3071178.3071229},
  \url{http://doi.acm.org/10.1145/3071178.3071229}

\bibitem{mnist-hyperneat}
Verbancsics, P., Harguess, J.: Image classification using generative neuro
  evolution for deep learning. In: 2015 IEEE Winter Conference on Applications
  of Computer Vision. pp. 488--493 (Jan 2015). \doi{10.1109/WACV.2015.71}

\end{thebibliography}

\end{document}